\documentclass[11pt,a4paper]{article}
\usepackage[hyperref]{acl2021}
\usepackage{times}
\usepackage{latexsym}

\usepackage{microtype}

\newcommand{\subfigimgtop}[3][,]{%
  \setbox1=\hbox{\includegraphics[#1]{#3}}
  \leavevmode\rlap{\usebox1}
  \rlap{\hspace*{350pt}\raisebox{\dimexpr\ht1-2.5\baselineskip}{#2}}
  \phantom{\usebox1}
}

\newcommand{\subfigimg}[3][,]{%
  \setbox1=\hbox{\includegraphics[#1]{#3}}
  \leavevmode\rlap{\usebox1}
  \rlap{\hspace*{350pt}\raisebox{\dimexpr\ht1-1.15\baselineskip}{#2}}
  \phantom{\usebox1}
}
\usepackage{graphicx}
\usepackage{tikz}
\usepackage{tikz-dependency}
\usetikzlibrary{shapes,decorations,arrows,calc,arrows.meta,fit,positioning}
\tikzset{
    -Latex,auto,node distance =0.5 cm and 1.5 cm,semithick,
    state/.style ={ellipse, draw, minimum width = 0.7 cm},
    point/.style = {circle, draw, inner sep=0.04cm,fill,node contents={}},
    bidirected/.style={Latex-Latex,dashed},
    el/.style = {inner sep=2pt, align=left, sloped}
}
\usepackage{caption}
\usepackage{subcaption}
\usepackage{multirow}

\aclfinalcopy 

\title{What if This Modified That?\\{S}yntactic Interventions via Counterfactual Embeddings}

\author{Mycal Tucker \\
  MIT\\
  \texttt{mycal@mit.edu} \\\And
  Peng Qian \\
  MIT \\
  \texttt{pqian@mit.edu} \\\And
  Roger P.\ Levy \\
  MIT \\
  \texttt{rplevy@mit.edu} \\}

\date{}

\begin{document}
\maketitle

\begin{abstract}
Neural language models exhibit impressive performance on a variety of tasks, but their internal reasoning may be difficult to understand.
Prior art aims to uncover meaningful properties within model representations via probes, but it is unclear how faithfully such probes portray information that the models actually use.
To overcome such limitations, we propose a technique, inspired by causal analysis, for generating counterfactual embeddings within models.
In experiments testing our technique, we produce evidence that suggests some BERT-based models use a tree-distance-like representation of syntax in downstream prediction tasks.
\end{abstract}

\section{Introduction}
Large neural models like BERT and GPT-3 have established a new state of the art in a variety of challenging linguistic tasks \cite{bertpaper,gpt3}.
These connectionist models, trained on large corpora in a largely unsupervised manner, learn to map words into numerical representations, or embeddings, that support language-reasoning tasks.
Fine-tuning these models on tasks like extractive question answering specializes these generic models into performant, task-specific models \cite{huggingface}.

In conjunction with the rise of these powerful neural models, researchers have investigated what the models have learned.
Probes, tools built to reveal properties of a trained model, are a favored approach  \cite{probeparser,conneau-etal-2018-cram}.
For example, \citet{hewitt2019structural} have uncovered compelling evidence that several models encode syntactic information in their embeddings.
That is, by passing embeddings through a trained probe, one may recover information about a sentence's syntax.

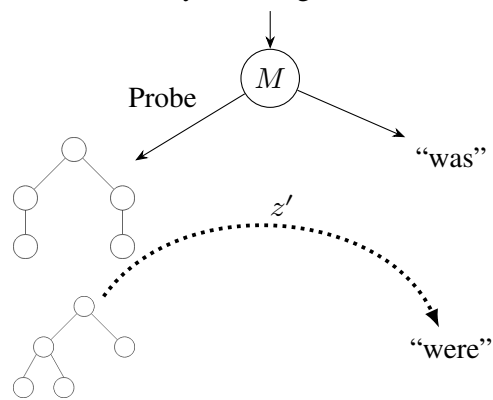
\begin{figure}[t]
    \centering
    \begin{tikzpicture}
        [
        basic/.style={draw, text centered},
        circ/.style={basic, circle, minimum size=2em, inner sep=1.5pt},
        rect/.style={basic, text width=1.5em, text height=1em, text depth=.5em},
        1 up 1 down/.style={basic, text width=1.5em, rectangle split, rectangle split horizontal=false, rectangle split parts=2},
        >={Stealth[]}
      ]
      \node at (0, 0) (input) {``I saw the boy and the girl \lbrack MASK\rbrack ~tall.''};
      \node[circ, below=of input] (M) {$M$};
    
    \node[inner sep=0pt, below left=of M] (P) {\includegraphics[width=0.1\textwidth]{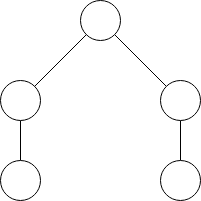}};
      \node[below right=of M] (A) {``was''};
    \node[inner sep=0pt, below=of P] (P2) {\includegraphics[width=0.1\textwidth]{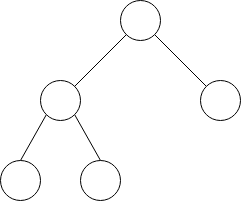}};
      \node[below=2cm of A] (A2) {``were''};

    \draw [->] (input) -- (M);
    \draw [->] (M) -- (P) node[near end, above=10pt] {Probe};
    \draw [->] (M) -- (A);
    \draw [-{Latex[length=2.5mm]},dotted,line width=0.5mm] (P2) to [bend left=60]  node [above, sloped]  (TextNode2) {$z'$} (A2);
    \end{tikzpicture}
    \caption{A language model, $M$, outputs predictions and a probe estimates properties from the model representation. We use probes to generate counterfactual representations, $z'$, based on syntactic manipulations, revealing reasoning within the model.
    }
    \label{fig:intro_sketch}
\end{figure}

Although these results are impressive, they fall short of clearly demonstrating what linguistic information the language models actually use.
Syntactic information is present in sentences; that embeddings also encode syntax does not imply that a model uses syntactic knowledge.

In order to truly query a model's understanding, one must use causal analysis.
Recently, several authors have done so by generating counterfactual data to test models \cite{Kaushik2020Learning,goyal2019explaining,elazar2020amnesic}.
They either create new input data or ablate parts of embeddings and study how model outputs change.
We extend this prior art via a new technique for generating counterfactual embeddings by using traditional probes to manipulate embeddings according to syntactic principles, as depicted in Figure~\ref{fig:intro_sketch}.
Because we conduct experiments with syntactically ambiguous inputs, we are able to measure how models respond to different valid parses of the same sentence instead of, for example, removing all syntactic information.
Thus, our technique uncovers not only what parts of its embeddings a model uses to represent syntax, but also how those parts influence downstream behavior.

Thus, in this work, we make two contributions.
First, we develop a gradient-based algorithm to generate counterfactual embeddings, informed by trained probes.
Second, in experiments using our technique, we find that the standard BERT model, trained on word-masking tasks, appears to leverage features of syntax in predicting masked words but that a BERT model fine-tuned for question-answering does not.
In addition, these experiments yield new data to inform the ongoing debate on probe design.\footnote{Code is available at \href{https://github.com/mycal-tucker/causal-probe}{https://github.com/mycal-tucker/causal-probe}}

\section{Related Work}

\subsection{Neural Language Model Probes}
Transformer-based models like GPT-3 and BERT have recently advanced the state of the art in numerous language-related problems \cite{gpt3,bertpaper,huggingface}.
These large models appear to learn meaningful representations of words and sentences, enabling high performance when fine-tuned for a specific task.

In conjunction with these models, probes have been developed to uncover what principles models have learned.
Such probes have been used in a wide variety of contexts, from image structure to syntax and semantics in language models (\citealp{alain2018understanding,conneau-etal-2018-cram,hewitt2019structural,coenen2019visualizing}, among others).
Our work uses two syntactic probes developed by \citet{hewitt2019structural} that map from model embeddings to predictions about word locations in a parse tree.
These probes are simple by design -- merely linear transformations -- in order to prevent the probes themselves from doing parsing.

Recent work directly addresses the topic of probe simplicity.
On the one hand, if probes are too expressive, they may reveal their own learning instead of a model's \cite{liu-etal-2019-linguistic,hewitt-liang-2019-designing}.
On the other hand, \citet{pimentel-etal-2020-information} argue from an information-theoretical perspective that more expressive probes are always preferable.

Our work differs from much prior art in probe design by leveraging causal analysis, which uses counterfactual data to test probes and models.
This provides direct evidence of whether a model uses the same features as a probe, allowing us to experiment beyond linear probes (and indeed, we found that more complex probes offered an advantage in some cases).

\subsection{Causal Analysis of Language Models}
Motivated by the limitations of traditional, correlative probes, researchers have recently turned to causal analysis to better understand language models.
\citet{goyal2019explaining} and \citet{Kaushik2020Learning} generate counterfactual inputs to language models, while \citet{vig2020causal} study individual neurons and attention heads to uncover gender biases in pre-trained networks.

Our work is most closely related to that of \citet{elazar2020amnesic}, who, as in this work, used probes to generate counterfactual embeddings within a network.
Their amnesiac counterfactuals are generated by suppressing features in embeddings that a probe uses.
In contrast, we use a continuous, gradient-based approach to generate counterfactuals, yielding insight into how features are used, as opposed to if they are used at all.

\section{Technical Approach}
\subsection{Problem Formulation}
We may characterize a transformer-based language model, $M$, trained on a specific task, as a function mapping from an input string, $s$, to an output $y$: $M(s) = y$.
In order to reveal embeddings for analysis by probes, we may decompose $M$ into two functions: $M_{k-}$ and $M_{k+}$.
$M_{k-}$ represents the first $k$ layers of the model; $M_{k+}$ represents the layers of $M$ after layer $k$; $M$ is the composition of these functions: $M = M_{k+} \circ M_{k-}$.
We label the embeddings output by $M_{k-}$ as $z_k$.
This decomposition of models to reveal internal embeddings mirrors the formulation for layer-specific probes \cite{hewitt2019structural}.
A probe may be defined as a function $f_p$ that maps from an embedding, $z_k$, to a predicted property $\hat{p}$ about the input, $s$: $f_p(M_{k-}(s)) = \hat{p}$.
(For the remainder of this paper, we focus on syntactic probes, but our reasoning may be extended to other linguistic properties.)

\begin{figure}
    \centering
    \begin{tikzpicture}
      [
        basic/.style={draw, text centered},
        circ/.style={basic, circle, minimum size=2em, inner sep=1.5pt},
        rect/.style={basic, text width=1.5em, text height=1em, text depth=.5em},
        1 up 1 down/.style={basic, text width=1.5em, rectangle split, rectangle split horizontal=false, rectangle split parts=2},
        >={Stealth[]}
      ]
      \node [circ] (z) {$z_k$};
      \node [circ, left=of z] (S) {$s$};
      \node [circ, above right=0.75cm of z] (zp) {$z_p$};
      \node [circ, below right=0.75cm of z] (zm) {$z_m$};
      \node [circ, right=of zp] (p) {$\hat{p}$};
      \node [circ, right=of zm] (y) {$y$};

      \draw[line width=0.25mm] [->] (S) -- (z) node [midway, fill=white] {$M_{k-}$};
      \draw[line width=0.25mm] [->] (z.45) -- (zp) node [midway, fill=white] {};
      \draw[line width=0.25mm] [->] (z.315) -- (zm) node [midway, fill=white] {};
      \draw[line width=0.4mm,  dotted] [->] (zp) -- (zm) node [midway, fill=white] {};
      \draw[line width=0.25mm] [->] (zm) -- (y) node [midway, fill=white] {$M_{k+}$};
      \draw[line width=0.25mm] [->] (zp) -- (p) node [midway, fill=white] {Probe};
    \end{tikzpicture}
    \caption{$M_{k-}$ yields a representation, $z_k$. $z_p$ and $z_m$ are subsets of the features of $z_k$ used by the probe and $M_{k+}$. We measured the causal link between $z_p$ and $z_m$.}
    \label{fig:causal_diagram}
\end{figure}
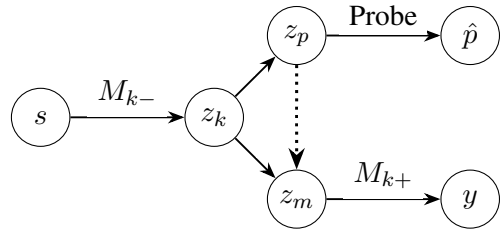

We may define two, potentially overlapping, subsets of the features of $z_k$ by considering different uses of $z_k$.
First, we may define $z_p$ as the features of $z_k$ that the probe uses in predicting $\hat{p}$ (for example, when using a linear probe, $z_p$ is the projection of $z_k$ onto the probe subspace).
Assuming good syntactic probe performance, $z_p$ is necessarily informative of the input's syntax.
We likewise defined $z_m$ as the features of $z_k$ that $M_{k+}$ uses in producing the model output.
These two, potentially overlapping, representations of $z_k$ are shown in Figure~\ref{fig:causal_diagram}, inspired by causal diagrams by \citet{bookofwhy}.
We seek to discover if there is a causal link between $z_p$ and $z_m$.

For some tasks, such a link should exist.
For example, a question-answering model's response to ``I shot the elephant wearing my pajamas. Who wore the pajamas?'' should depend upon the inferred sentence syntax (e.g., if the probe predicts that ``wearing my pajamas'' modifies ``the elephant,'' the model should output ``the elephant'').
Thus, the probe and model outputs should ``agree'' according to syntactic principles.
Furthermore, if a causal link between $z_p$ and $z_m$ exists, changing $z$ to produce a new prediction of syntax should change the model output to agree with the probe (e.g., if the probe predicts that ``wearing my pajamas'' now modifies ``I,'' the model should now output ``I'').
In this work, therefore, we study whether a link between $z_p$ and $z_m$ exists and, if it does, to what extent it corresponds with linguistic principles.

\subsection{Generating Counterfactual Embeddings via Gradient Descent}
To study such a link, we must generate counterfactual embeddings, $z'$, that modify probe outputs, starting from normal embeddings $z_k$.
We borrow the term ``counterfactual'' from causal literature because $z'$ represents what $z_k$ would have been if $z_p$ had been different \cite{bookofwhy}.
We were particularly interested in finding $z'$ that changed both probe and model outputs; if $z'$ only changed probe outputs, that could indicate that the probe was over-interpreting model embeddings (e.g., acting as a parser instead of a probe).\footnote{We did not study $z'$ that only modified the model outputs, although this could be a promising avenue for future work.}


We developed a gradient-based method to generate $z'$ that changed the probe output.
We assumed that, given the probe function, $f_p$, a loss, $L$, and the correct property value (e.g., parse), $p$, one could compute the gradient of the loss with respect to the probe inputs: $\nabla_{z'} L(f_p(z'), p)$.
Neural network probes obey such differentiability assumptions.

Given $z_k$ and $p$, we constructed a counterfactual embedding , $z'$, by initializing $z'$ as a $z_k$ generated by the model and updating $z'$ via gradient descent of the loss.
Updating $z'$ may be terminated based on various stopping criteria (e.g., local optimality, loss below a threshold, etc.), yielding the final counterfactual $z'$.
Assuming non-zero gradients, this technique produces $z'$s that, by design, change the probe outputs.
In experiments, we studied how $z'$s changed model outputs when passed through $M_{k+}$.

Although our technique bears some resemblance to gradient-based adversarial attacks \citep{advimage}, it may more broadly be thought of as guided search in a latent space.
Adversarial images are often characterized by changes that are imperceptible to humans but change model behaviors to be incorrect.
In contrast, we seek to find embeddings that change both probe and language model outputs.
Furthermore, by design, we use syntactically ambiguous sentences in experiments and generate counterfactuals according to valid parses.
Thus, unlike adversarial attacks on images that seek to switch model classification to an incorrect class, we merely guide embeddings among a set of valid interpretations.
Lastly, even uncovering instances of embeddings that change probe outputs but not the model's is important, as it indicates a misalignment of probe and model reasoning.

\begin{table*}[t]
    \centering
    \begin{tabular}{c|c|l|l}
    Model & Corpus & Parse & Example Input \\
    \hline
    \multirow{4}{*}{Mask}     & \multirow{2}{*}{Coord.} &  Plur. & The woman saw (the boy and the dog) [MASK] falling.\\
    & & Sing. & (The woman saw the boy) and (the dog [MASK] falling.)\\
    \cline{2-4}
    & \multirow{2}{*}{NP/Z} &  Adv. & When the dog scratched (the vet [MASK] ran.)\\
    & & Noun & (When the dog scratched the vet) [MASK] ran.\\
    \hline
    \multirow{4}{*}{QA}     & \multirow{2}{*}{RC} &  Conj. & The ((smart women and rich men) who were desperate) bribed the judge.\\
    & & NP2 & The (smart women) and (rich men who were desperate) bribed the judge.\\
    \cline{2-4}
    & \multirow{2}{*}{NP/VP} & VP & The girl saw (the boy) with the telescope.\\
    & & NP2 & The girl saw (the boy with the telescope.)\\
    \end{tabular}
    \caption{Experiment design for different language models and test corpora, with illustrative sentences, decorated with auxiliary parentheses to reveal structure. The parentheses were not included in the actual corpora.}
    \label{tab:experiments}
\end{table*}

\section{Experiments}
In the previous section, we proposed a technique for generating counterfactual embeddings; here, we detailed the experiments we conducted to measure the effect of using such embeddings.
Inputs to our technique included the base language models, probes, test sentences, and different ground-truth parses to generate the counterfactual embeddings.

\subsection{Model Tasks}
We tested our technique on two BERT models trained on different tasks: masked word prediction and extractive question answering.

In the masked word prediction task, a model is given a sentence, $S$, comprising words ($s_0, s_1, ...$ \texttt{[MASK]}$,..., s_n$) and must predict the word at the location marked by \texttt{[MASK]}.
For example, given a sentence, [`The', `children', `went', `out', `to', \texttt{[MASK]}, `.'], a correct answer might be `play.'
We used huggingface's ``bert-large-uncased-whole-word-masking'' model, which was trained on masked word and next-sentence prediction, and referred to it as the ``Mask'' model \cite{huggingface}.

Extractive question answering is framed by \citet{huggingface} as follows:
given a sentence, $S$, comprising word tokens ($s_0, s_1,... s_n$) and a question, identify the start and end tokens ($s_i, s_j; 0 \leq i \leq j \leq n$) denoting a contiguous stretch of the sentence that answers the question.
For example, given the sentence [`I', `ate', `two', `apples', `.'] and the question ``How many apples did I eat?," a correct answer could be [2, 2] (``two'') or [2, 3] (``two apples'').
We used huggingface's ``BertForQuestionAnswering,'' already fine-tuned on the SQuAD dataset, and referred to the model as QA \cite{huggingface,squad}.

\subsection{Probes}
Our technique for generating counterfactual embeddings depended on probes, so we used four different syntactic probes drawn from prior art and our own design.

The depth probe from \citet{hewitt2019structural} maps from embeddings to predictions over words' depths in a sentence's parse tree.
The distance probe, given a pair of words, predicts the distance between the words in the parse tree (i.e., how many edges must be traversed).
Both probes consist of a linear transformation from embedding to prediction.

We further implemented ``deep'' versions of the distance probe by creating two- and three-layer, non-linear probes trained on the distance task.
These models used ReLU activations, with hidden dimension 1024, but otherwise used the same input and output format as the linear distance probe.
(Experiments conducted with ``deep'' versions of the linear depth probe produced similar results to those of the normal depth probe and are therefore omitted.)

\subsection{Evaluation Corpora}
We used four corpora for evaluating the Mask and QA models, as summarized in Table~\ref{tab:experiments}.

\subsubsection{Mask Test Corpora}
For the Mask model, we used two test suites composed of sentences whose structural ambiguity was resolved by a masked word.

The first corpus, dubbed ``Coordination,'' comprised sentences that took the form ``The NN1 VERB the NN2 and the NN3 [MASK] ADJ.''
Such sentences may be interpreted in at least two ways by inserting either ``was'' or ``were'' in the masked location.
The former reflects a conjunction of clauses (e.g., ``The woman saw the boy and the dog was falling.''), whereas the latter reflects a conjunction of noun phrases (e.g., ``The woman saw the boy and the dog were falling.'')
Sentences were generated through combinations of NN1 [man, woman, child], VERB [saw, feared, heard], NN2 [boy, building, cat], NN3 [dog, girl, truck], and ADJ [tall, falling, orange], yielding 243 sentences, each with two parse trees dubbed ``singular'' or ``plural,'' depending on the grammatical verb type.

The second corpus, dubbed the NP/Z corpus, was inspired by classic psycholinguistic studies of the garden-pathing effect in online sentence processing \citep{frazier-rayner:1982,tabor2004evidence}.
Each sentence in the corpus took the form ``When the NN1 VERB1 the NN2 [MASK] VERB2.''
Without knowing the masked word, it is unclear if NN2 is the object of the subordinate clause or the subject of the main clause.
For example, in the sentence ``When the dog scratched the vet [MASK] ran,'' either an adverb (e.g., ``immediately'') or a noun (e.g., ``she'') would be permitted but correspond to different parses.
We created such parse trees and dubbed the first type ``Adv.'' and the second type ``Noun.''
We used the 24 sentences from \citet{tabor2004evidence} that fit our template, and supplemented the dataset with 36 sentences of our own, generated by iterating over all combinations of NN1 [dog, child], NN2 [vet, boy, girl], VERB1 [scratched, bit], and VERB2 [ran, screamed, smiled].
(Augmenting the dataset was needed to increase the statistical analysis power, and plotting the 24 and 36 sentences separately established that they produced similar results.)

\subsubsection{QA Test Corpora}
For the QA model, we created two test suites.
First, the ``RC'' corpus used sentences composed of a conjunction of nouns modified by a relative clause.
All sentences took the form ``The ADJ1 NN1 and ADJ2 NN2 who were ADJ3 VERB the NN3. Who was ADJ3?''
For example, one sentence was ``The smart women and rich men who were desperate bribed the judge. Who was desperate?''
By construction, it was unclear if the relative clause modified the conjunction of the first and second noun phrases (The ADJ1 NN1 and ADJ2 NN2) or merely the second noun phrase (ADJ2 NN2).
For each sentence, we generated two parses: ``Conj. Parse'' and ``NP2 Parse,'' corresponding to the former and latter.
We generated sentences by iterating over all combinations of values for ADJ1 [smart, rich, tall, poor], NN1 [men, women], ADJ2 [smart, rich, tall, poor], NN2 [men, women], ADJ3 [corrupt, desperate], VERB [bribed, paid], and NN3 [politician, judge], excluding sentences in which NN1 and NN2 or ADJ1 and ADJ2 were the same.
This produced 192 sentences, each with two parses.

Lastly, the ``NP/VP'' corpus used sentences with ambiguous prepositional phrase attachment.
Inspired by sentences like ``The girl saw the boy with the telescope,'' we generated inputs with the template ``The NN1 VERB the NN2 with the NN3. Who had the NN3?''
We iterated through combinations of NN1 [man, woman, child], NN2 [man, woman, boy, girl, stranger, dog], and VERB-NN3 pairs [saw-telescope, poked-stick, thanked-letter, fought-knife, dressed-hat, indicated-ruler, kicked-shoe, welcomed-gift, buried-shovel], removing duplicate NN1 and NN2, yielding 144 inputs.
Each input used two parses indicating the prepositional phrase modifying VP or NP2 (``the'' and NN2).

\subsection{Generating Embeddings}
For all models, probes, and parses trees for each sentence, we generated counterfactual embeddings by initializing a counterfactual embedding, $z'$, as the original model embedding for the input sentence, $z_k$, and running an Adam optimizer, with learning rate $0.0001$, to minimize the probe loss (using a particular probe and parse tree) \cite{adam}.
Recall that the optimizer updated $z'$ rather than the probe parameters.

The optimizer used a patience value of 5000: it continued updating $z'$ until the probe loss failed to improve for 5000 consecutive gradient updates.
Using a patience-based termination condition (as opposed to setting a loss threshold or maximum number of updates, for example) was task-agnostic and seemed to be robust to a wide range of patience values.
Brief experimentation with patience values from 50 to 5000 yielded similar results.
On a Linux desktop with an Nvidia GEForce RTX 2080 graphics card, generating a single counterfactual took less than 1 minute, and the process was easily parallelized to batches of 80 embeddings, reducing the mean computation time to under one second.

For both the QA and Mask models, we trained all probe types (depth, distance, 2-layer dist, and 3-layer dist) on each of the model's 25 layers.
We used 5000 entries from the Penn Treebank (PTB) for training, with the standard validation and test sets of nearly 4000 entries used for early stopping and evaluation, respectively \cite{ptb}.

\begin{figure*}[!htb]
    \centering
    \begin{subfigure}[b]{0.48\textwidth}
        \centering
        \includegraphics[width=0.95\textwidth, trim={0.3cm 0.5cm 0.5cm 0.25cm}, clip=True]{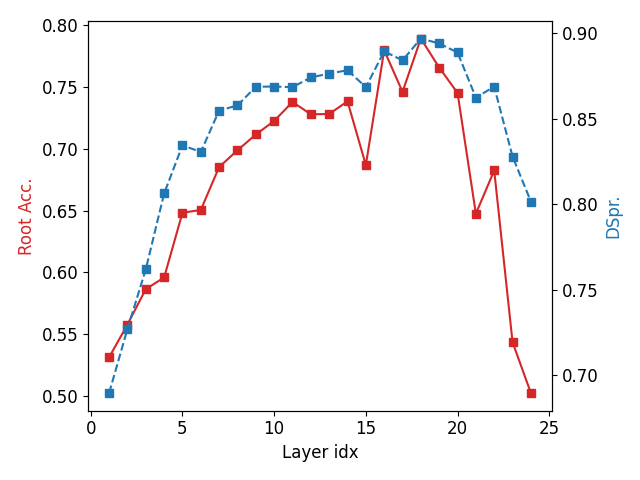}
        \caption{Depth Probe}
    \end{subfigure}
    \hfill
    \begin{subfigure}[b]{0.48\textwidth}
        \centering
        \includegraphics[trim={0.3cm 0.5cm 0.5cm 0.25cm}, clip=True, width=0.95\textwidth]{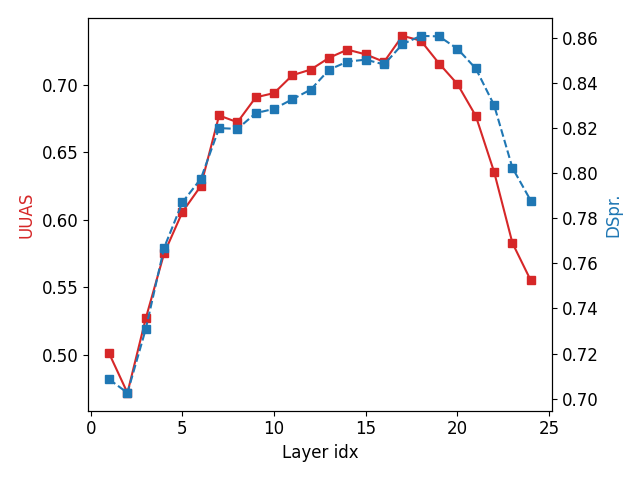}
        \caption{Dist Probe}
    \end{subfigure}
    \begin{subfigure}[b]{0.48\textwidth}
        \centering
        \includegraphics[width=0.95\textwidth, trim={0.3cm 0.5cm 0.5cm 0.25cm}, clip=True]{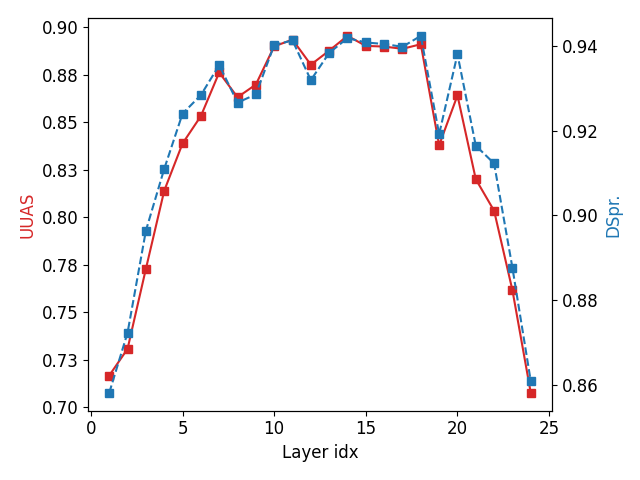}
        \caption{2-layer Dist Probe}
    \end{subfigure}
    \hfill
    \begin{subfigure}[b]{0.48\textwidth}
        \centering
        \includegraphics[width=0.95\textwidth, trim={0.3cm 0.5cm 0.5cm 0.25cm}, clip=True]{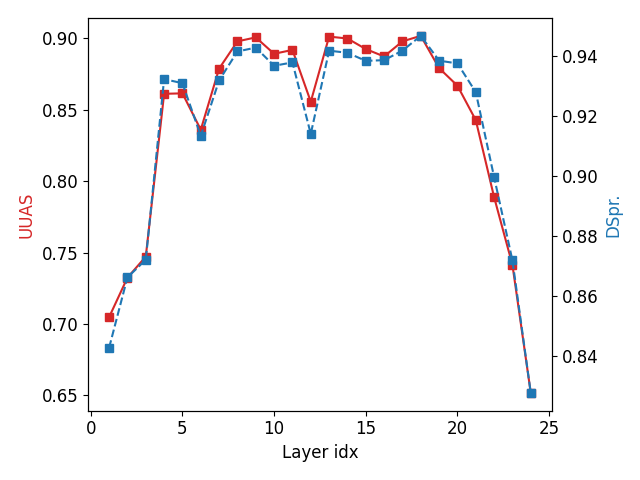}
        \caption{3-layer Dist Probe}
    \end{subfigure}
    \caption{All trained probes for the QA model exhibited high performance on the PTB corpus.}
    \label{fig:probeperfs}
\end{figure*}

\subsection{Metrics}
\label{sec:metrics}
We used two sets of metrics in our experiments.
First, we measured probe performance using the Root Accuracy, UUAS, and Spearman Coefficient metrics used by \citet{hewitt2019structural} and refer to their work for details.
Intuitively, these metrics captured how accurately the probes predicted aspects of syntactic structure from embeddings.

\begin{figure*}[!htb]
    \centering
    Mask Model Prediction for ``as the author wrote the book [MASK] grew.''
    \begin{subfigure}[b]{\textwidth}
        \includegraphics[trim={0cm 0.95cm 0cm 0.85cm}, clip=true, width=0.97\textwidth]{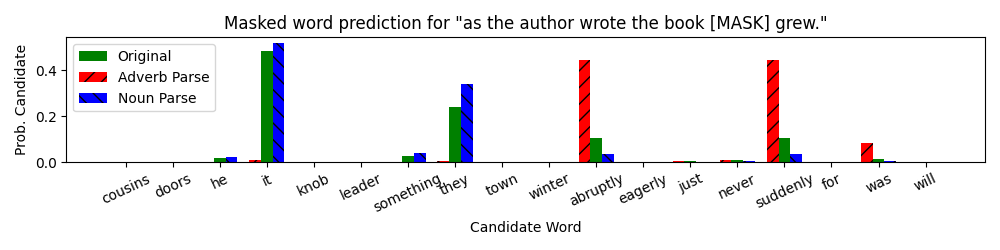}
    \end{subfigure}
    ~
    \begin{subfigure}[b]{\textwidth}
        \centering
        \begin{dependency}[hide label, edge unit distance=.5ex]
            \begin{deptext}[column sep=0.05cm]
            as\& the\& author\& wrote\& the\& book\& \lbrack MASK \rbrack \& grew\& . \\
            \end{deptext}
            \depedge[edge style={red!60!}, edge above]{7}{8}{.}
            \depedge[edge style={red!60!}, edge above]{2}{3}{.}
            \depedge[edge style={red!60!, ultra thick}, edge above]{6}{8}{.}
            \depedge[edge style={red!60!}, edge above]{3}{4}{.}
            \depedge[edge style={red!60!}, edge above]{1}{4}{.}
            \depedge[edge style={red!60!}, edge above]{4}{8}{.}
            \depedge[edge style={red!60!}, edge above]{5}{6}{.}
            
            \depedge[edge style={blue!60!, ultra thick}, edge below]{4}{6}{.}
            \depedge[edge style={blue!60!}, edge below]{7}{8}{.}
            \depedge[edge style={blue!60!}, edge below]{3}{4}{.}
            \depedge[edge style={blue!60!}, edge below]{2}{3}{.}
            \depedge[edge style={blue!60!}, edge below]{1}{4}{.}
            \depedge[edge style={blue!60!}, edge below]{5}{6}{.}
            \depedge[edge style={blue!60!}, edge below]{4}{8}{.}
        \end{dependency}
    \end{subfigure}
    \caption{Given a sentence from the NP/Z corpus, the Mask model originally predicted ``it'' or ``they,'' but using counterfactuals from the $5^{th}$ layer 3-layer dist probe changed predictions to favor nouns (cousins - winter) or adverbs (abruptly - suddenly). Visualizing the word dependencies revealed that the Adverb parse (top, red) and Noun parse (bottom, blue) induced different dependencies (differences in bold), as expected.}
    \label{fig:example_dist_cloze}
\end{figure*}

Second, we measured changes in model outputs when using counterfactual embeddings.
The Mask model produced a probability distribution over more than 30,000 possible words for the masked location, but we restricted our attention to only a subset of those words, dubbed ``candidates.''
(We normalized predictions among the set of candidates, producing a proper probability distribution.)
In the Coordination corpus, we used 5 candidates: [``was,'' ``is,'' ``were,'' ``are,'' ``as''].
In the NP/Z corpus, we generated the set of candidates by collecting the most likely predictions over the corpus, using both original and counterfactual embeddings.
This set of 18 words is shown in the $x$-axis of Figure~\ref{fig:example_dist_cloze}.
For both corpora, we partitioned the candidates into two sets, depending upon which parse they implied, and measured the sum of the probabilities of words in each set.
If counterfactual embeddings caused the models to change the type of word they predicted, we would expect to see a change in these sums.

For the QA model, we similarly measured changes in probabilities among sets of words, but in this case we focused on the predicted start location of the answer.
Recall that the QA model produced two distributions over words, indicating its predictions over where the answer started and ended.
Consider an example input, drawn from the RC corpus: ``The smart women and rich men who were desperate bribed the politician. Who was desperate?''
Two reasonable answers might be ``The smart women and rich men'' or ``rich men,'' corresponding to QA outputs with identical end words, but differing start words.
We therefore created two partitions of starting words to consider: those belonging to the first noun phrase (``The smart women'') or the second noun phrase (``rich men'').
We then measured the summed start probabilities of words in each partition.
We did not normalize these probabilities, as the QA model rarely predicted start words outside these two partitions with more than 1\% probability.

In all experiments, we employed one-sided Wilcox Signed-Rank tests, non-parametric tests for pairwise data, when determining the significance at $(p < 0.01)$.
The parses were viewed as ``treatments'' for the same embedding.
We compared the effect of using counterfactual instead of original embeddings, as well as the effect of using different parses to generate counterfactual embeddings.

\begin{figure*}[!htb]
    \centering
        Mask Model Coord. Corpus Likelihood of Plural Candidates by Layer
    \begin{subfigure}[b]{0.98\textwidth}
        \includegraphics[width=0.97\linewidth]{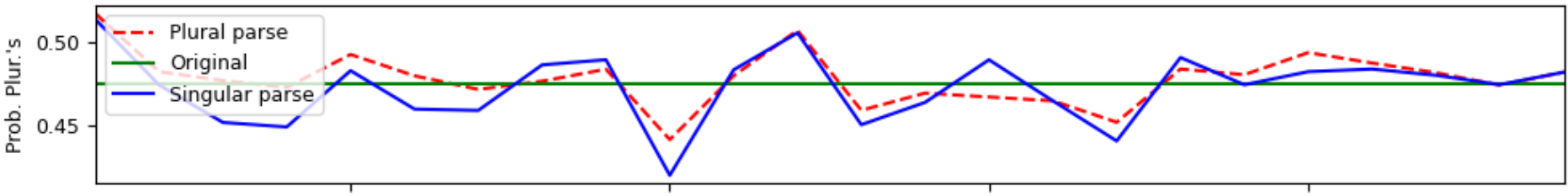}
    \end{subfigure}
    \begin{subfigure}[b]{\textwidth}
        \includegraphics[trim={0cm 0.35cm 0cm 1.05cm}, clip=true, width=0.97\textwidth]{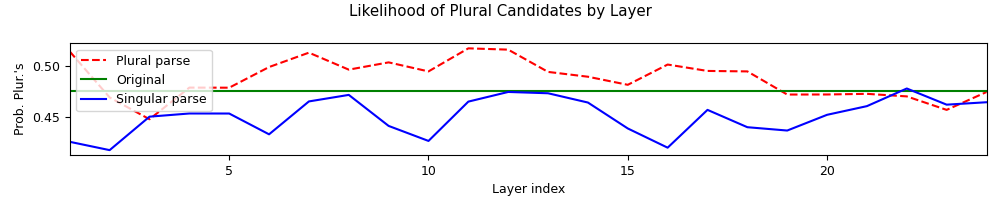}
    \end{subfigure}
    \caption{Mean probability of plural candidates using the depth probe (top) or the 3-layer dist probe (bottom), using original or counterfactual embeddings, in the Coordination corpus. Using a parse that implied plural words increased the probability of plural words when using the 3-layer dist probe.}
    \label{fig:cloze_coord}
\end{figure*}

\begin{figure*}[!htb]
    \centering
    Mask Model NP/Z Corpus Likelihood of Adverb Candidates by Layer
    \begin{subfigure}[b]{\textwidth}
        \includegraphics[trim={0cm 1.2cm 0cm 1.05cm}, clip=true, width=0.97\textwidth]{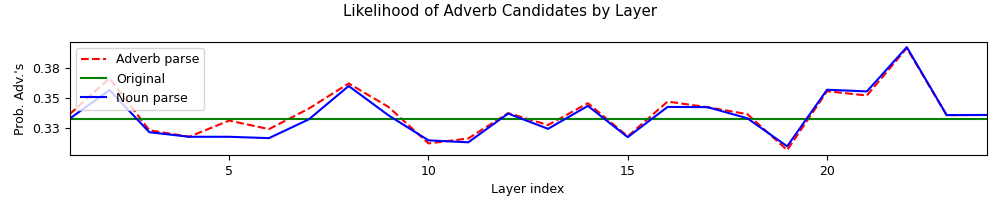}
    \end{subfigure}
    ~
    \begin{subfigure}[b]{\textwidth}
        \includegraphics[trim={0cm 0.35cm 0cm 1.05cm}, clip=true, width=0.97\textwidth]{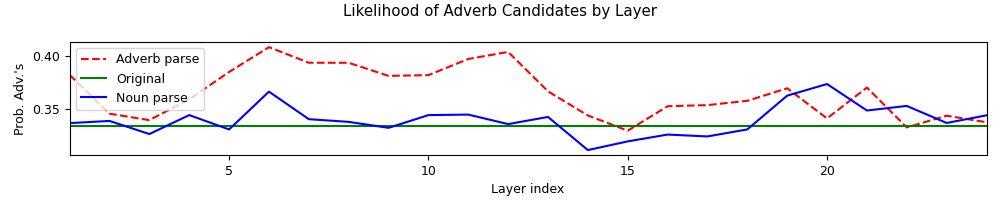}
    \end{subfigure}
    \caption{Mean probability of adverb candidates in the NP/Z corpus, using original and counterfactual embeddings generated by the depth (top) and 3-layer dist probes (bottom).}
    \label{fig:digging_in}
\end{figure*}

\section{Results}
Our results indicated that our probes performed well, as evaluated by performance metrics from prior art.
However, we found that only some combinations of probe types and BERT models generated counterfactuals that altered the model's outputs according to syntactic principles.

\subsection{Probe Performances}
Measured on the PTB test set, the probe performance metrics confirmed that the probes predicted aspects of syntactic structure well \cite{ptb}.
Plots of performance, similar to those by \citet{hewitt2019structural}, for probes trained on QA model embeddings are included in Figure~\ref{fig:probeperfs}.\footnote{All probe metrics are plotted in the appendix.}
For both models and all probe types, we found that the probes were able to achieve high performance, indicating that both the Mask and QA models encoded syntactic information in their embeddings.

We also observed the unsurprising trend that multi-layered, non-linear distance probes outperformed the linear distance probe.
This raised the question, if different probes exhibited different performance for the same model, which probe should be used to deduce model behavior?
Injecting counterfactual embeddings generated by different probes helped us answer this question.

\subsection{Mask Counterfactual Results}
Next, we found that using the distance-based probes to generate counterfactual embeddings in the Mask model consistently produced the desired effect by shifting the model's prediction of the masked word according to syntactic principles, and that the multi-layer distance probes performed better than the linear probe.

We plotted the mean effect of counterfactual embeddings for the Coordination and NP/Z corpora in Figures~\ref{fig:cloze_coord} and \ref{fig:digging_in}, respectively.\footnote{Plots for the effects of counterfactuals for all probes, models, and test corpora were included in the appendix.}
Each plot depicts the mean prediction likelihood of one of the partitions of candidates (plural for Coord. corpus, adverbs for NP/Z), using original or counterfactual embeddings.
Figure~\ref{fig:cloze_coord} shows results using the depth and 3-layer distance probes in the Coord. corpus: the depth probe failed to produced consistent changes in word probabilities, but embeddings generated by the 3-layer dist probe did exhibit the desired effect.
The change in probability of plural words when using the plural parse was significantly positive for layers 6 through 14 (among others) and greater than the change when using the singular parse for layers for 4 through 21.

\begin{figure*}[!htb]
    \centering
    QA Model Likelihood of NP1 Start by Layer
    \begin{subfigure}[b]{\textwidth}
        \centering
        \includegraphics[width=0.98\textwidth]{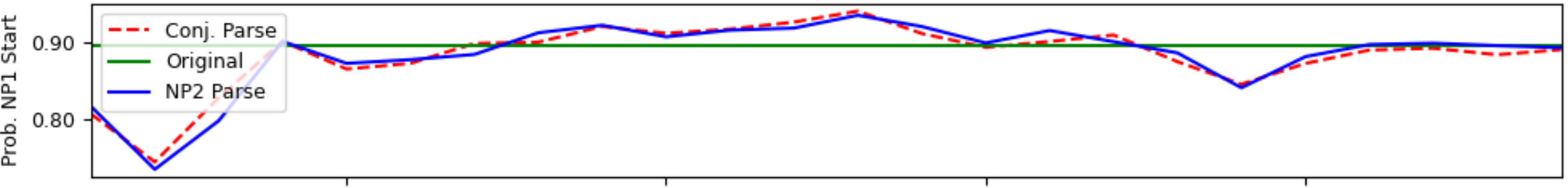}
    \end{subfigure}
    ~
    \begin{subfigure}[b]{\textwidth}
        \includegraphics[trim={0cm 0.35cm 0cm 1.05cm}, clip=true, width=\textwidth]{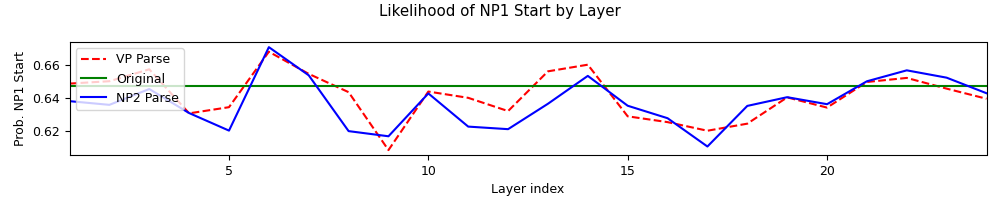}
    \end{subfigure}
    \caption{Mean effects of using counterfactual updates from the 3-layer dist probe on the QA model for the RC (top) and NP/VP (bottom) corpora.}
    \label{fig:qa_graphs}
\end{figure*}

Similar results were observed using the 3-layer distance probe for the NP/Z corpus, as shown in Figure~\ref{fig:digging_in}.
The net increase in probability for adverbs when using the adverb parse was significantly greater than when using the NP2 parse for layers 5 through 19 and was positive for layers 4 through 13.

We examined an example sentence from the NP/Z corpus in Figure~\ref{fig:example_dist_cloze} in greater depth.
The 18 words displayed along the $x$ axis were the candidate words whose probabilities we calculated in the NP/Z corpus.
As expected, using the Adv. parse increased the likelihood of adverbs like ``suddenly,'' while using the Noun parse increased the likelihood of nouns like ``it'' or ``they.''
Lastly, the bottom part of Figure~\ref{fig:example_dist_cloze} shows the dependency trees for the counterfactuals generated for each parse (see \citet{hewitt2019structural} for details on creating such trees).
These trees reflected the dependencies of the parses that generated the counterfactuals, indicating that our technique changed embeddings in the way we intended.


Together, the results from both corpora, revealed that distance-, but not depth-, based probes elicited the desired response from the Mask model, which suggests that it leverages a distance-based representation of syntax in its reasoning.

\subsection{QA Counterfactual Results}

Lastly, we examined the effect of using counterfactual embeddings in the QA model.
Compared to the Mask model, we found smaller and less consistent results, suggesting that the QA model may not use syntax.

Taking the mean across sentences in the corpus, we plotted the mean starting probabilities of words in each sentence's first noun phrase (as explained earlier in Section~\ref{sec:metrics}).
These values reflect whether the model predicted NP1 should be included in the answer (e.g., ``The smart women and rich men'' instead of merely ``rich men'').
We plotted the results for the 3-layer dist probe, the best-performing probe for the Mask model, on both QA corpora in Figure~\ref{fig:qa_graphs}.
In both plots, the choice of layer in which counterfactuals were inserted had a greater effect than which parse was used to generate the counterfactuals -- a sign of poor performance.
Depth and other distance probes performed no better.

Visualizing dependency trees for QA embeddings revealed that the counterfactual embeddings induced the correct structure, indicating that the QA model simply did not use such structure in downstream predictions.
Furthermore, given the success of our probes and technique with the Mask model, these poor results for the QA model suggest (but admittedly cannot definitely prove) that it may not have learned to use the syntactic information detected by the probes.
This theory is consistent with prior art that finds that fine-tuning on specific tasks, as was done for the QA model, worsens the alignment between model and human representations of language \cite{Gauthier2019LinkingAA}.

\section{Conclusion}
In this work, we proposed and evaluated a new technique for producing counterfactual embeddings that tested syntactic understanding of models and probes.
On the one hand, we uncovered clear evidence supporting a causal link between a distance-based representation of syntax and the outputs of a masked-word model.
On the other hand, depth-based manipulations of embeddings had little effect, and we found no evidence that the BERT model finetuned on question-answering uses the syntactic information used by probes.

Our work is merely an initial step in the direction of causal analysis of language models.
Developing new probes, backed by causal evidence, could increase our understanding of such models.
In particular, our findings that multi-layered probes outperformed linear probes indicate that the prior guidance of simpler probes being preferable may be misleading.
Furthermore, as the discrepancy between distance- and depth-based probes revealed, developing a large suite of probe types that focus on different features may be necessary to reveal a model's reasoning.
In tandem with probe development, more sophisticated counterfactual generation techniques than our gradient-based method could produce more interesting counterfactuals for evaluation.

\section*{Acknowledgments}
We thank the reviewers for their thoughtful comments, in particular regarding adversarial attacks.
We thank Professors Julie Shah and Jacob Andreas for ongoing discussions and guidance.
Lastly, we thank John Hewitt and Christopher Manning for releasing high-quality, reproducible code, enabling us to rapidly build upon their syntactic probe codebase.

RPL gratefully acknowledges support from the MIT--IBM Artificial Intelligence Laboratory and MIT's Quest for Intelligence.

\bibliographystyle{acl_natbib}
\bibliography{anthology,acl2021}

\clearpage
\section*{Appendix: Complete Performance Plots}
In this appendix, we included additional figures that we were unable to include within the main paper limits.

\begin{figure*}
    \centering
    \begin{subfigure}[b]{0.45\textwidth}
        \centering
        \includegraphics[width=\textwidth, trim={0.0cm 0cm 0.0cm 0cm}, clip=True]{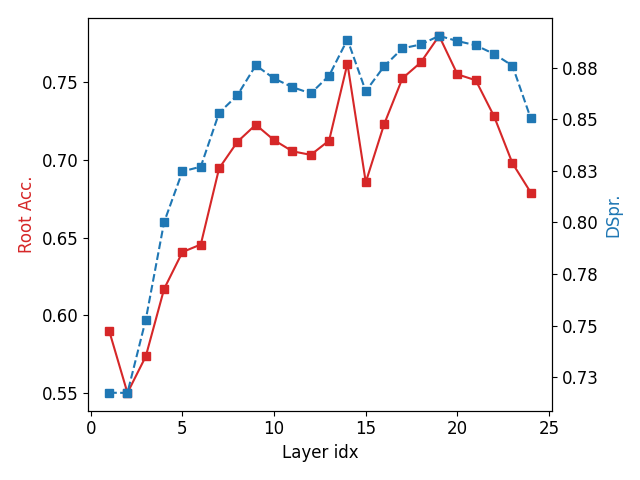}
        \caption{Mask Model Depth Probe}
    \end{subfigure}
    \hfill
    \begin{subfigure}[b]{0.45\textwidth}
        \includegraphics[width=\textwidth, trim={0.0cm 0cm 0.0cm 0cm}, clip=True]{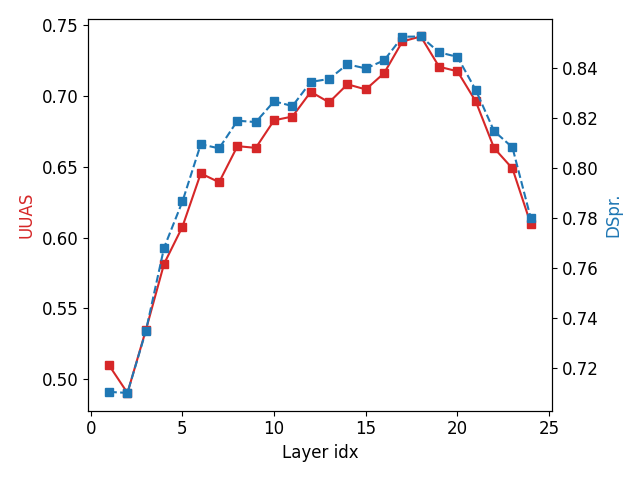}
        \caption{Mask Model Dist Probe}
    \end{subfigure}
    ~
    \begin{subfigure}[b]{0.45\textwidth}
        \centering
        \includegraphics[width=\textwidth, trim={0.0cm 0cm 0.0cm 0cm}, clip=True]{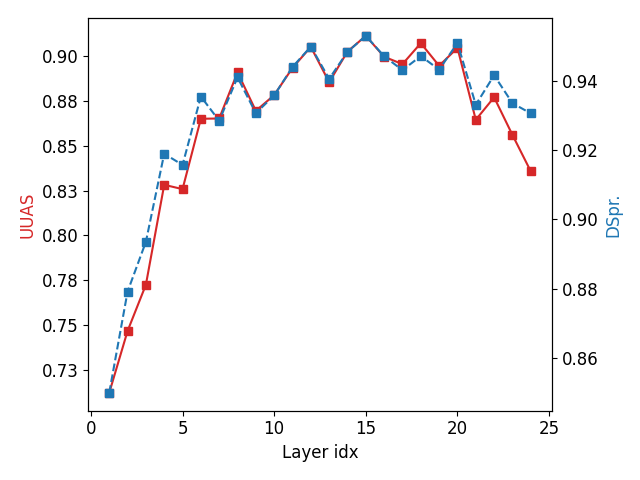}
        \caption{Mask Model 2-Layer Dist Probe}
    \end{subfigure}
    \hfill
    \begin{subfigure}[b]{0.45\textwidth}
        \includegraphics[width=\textwidth, trim={0.0cm 0cm 0.0cm 0cm}, clip=True]{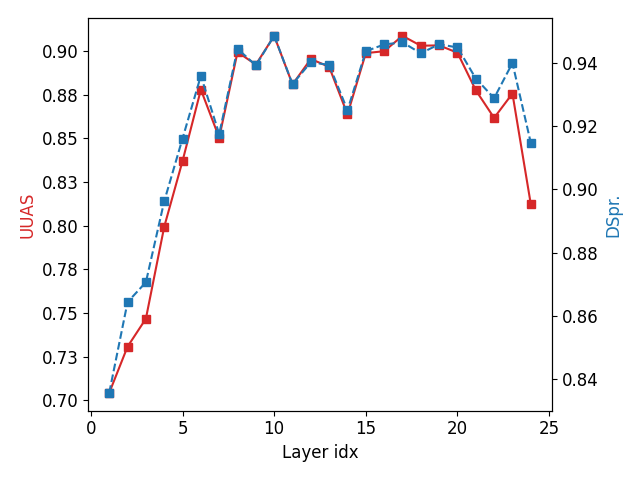}
        \caption{Mask Model 3-Layer Dist Probe}
    \end{subfigure}
    ~
    \begin{subfigure}[b]{0.45\textwidth}
        \centering
        \includegraphics[width=\textwidth, trim={0.0cm 0cm 0.0cm 0cm}, clip=True]{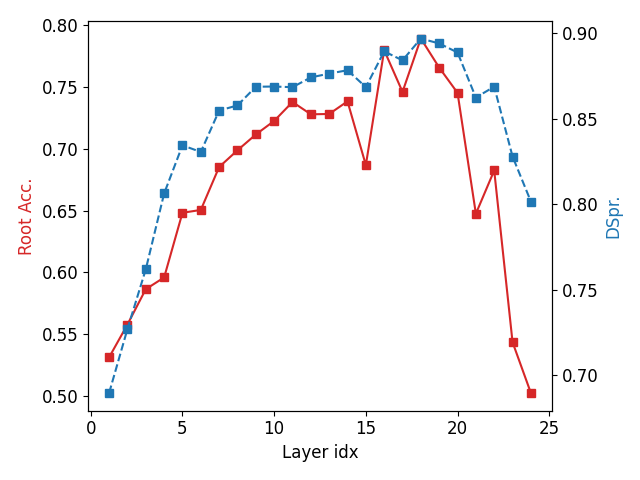}
        \caption{QA Model Depth Probe}
    \end{subfigure}
    \hfill
    \begin{subfigure}[b]{0.45\textwidth}
        \includegraphics[width=\textwidth, trim={0.0cm 0cm 0.0cm 0cm}, clip=True]{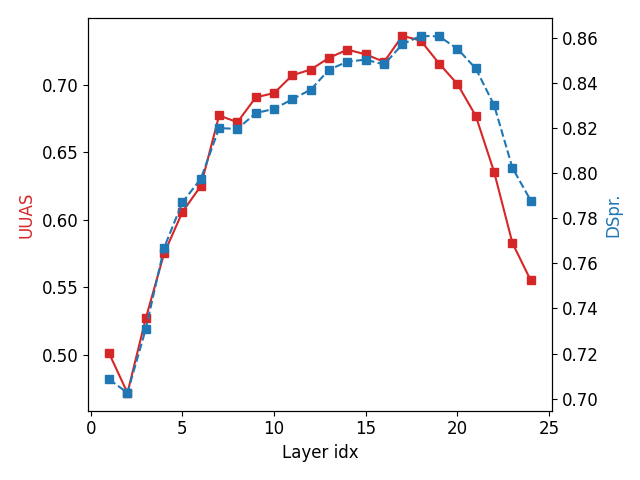}
        \caption{QA Model Dist Probe}
    \end{subfigure}
    ~
    \begin{subfigure}[b]{0.45\textwidth}
        \centering
        \includegraphics[width=\textwidth, trim={0.0cm 0cm 0.0cm 0cm}, clip=True]{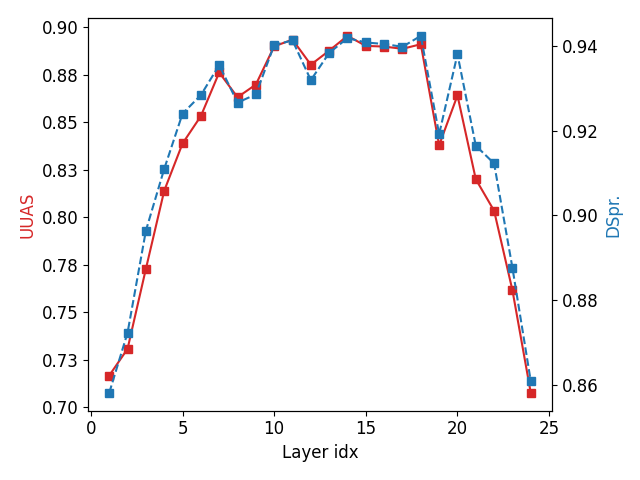}
        \caption{QA Model 2-Layer Dist Probe}
    \end{subfigure}
    \hfill
    \begin{subfigure}[b]{0.45\textwidth}
        \includegraphics[width=\textwidth, trim={0.0cm 0cm 0.0cm 0cm}, clip=True]{supplementaryfigures/probe_perf/dist3_probe_perf.png}
        \caption{QA Model 3-Layer Dist Probe}
    \end{subfigure}
    \caption{Probe performances for the Mask and QA models. Note the changed y axes, demonstrating improved performance for the multi-layer distance probes.}
    \label{appfig:probe_perfs}
\end{figure*}

First, we depicted the probe performance characteristics for the 4 probes types we used in all our experiments: the depth, dist, 2-layer dist, and 3-layer dist probes.
Each type of probe was trained for both the QA and Mask models.
Evaluation of these probes was plotted in Figure~\ref{appfig:probe_perfs}.

Next, we reported the effect of counterfactual embeddings generated for each model, corpus, and probe type.
Given the 4-page limit for the appendix, further plots breaking down the NP/Z corpus, for example, or depicting performance for multi-layered depth probes were not included.
These plots merely confirmed trends already present in the data: that depth-based probes did not produce useful counterfactuals, and that the curated and automatically-generated sentences that formed the full NP/Z corpus yielded similar results.

\begin{figure*}[!htb]
    \centering
    Mask Model Likelihood of Plural Candidates by Layer in Coordination Corpus
    \begin{subfigure}[b]{1.0\textwidth}
        \subfigimg[trim={0cm 1.3cm 0cm 1cm}, clip=true, width=\textwidth, height=2cm]{Depth}{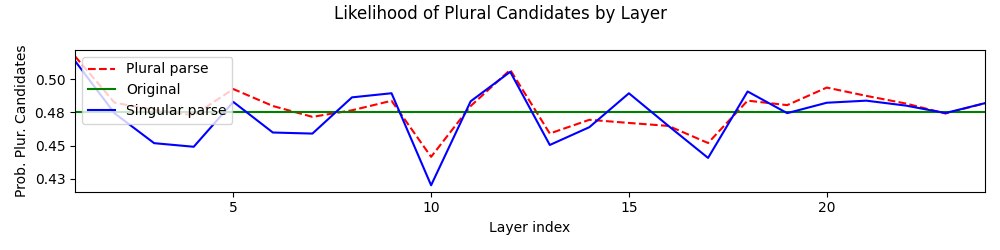}
    \end{subfigure}
    ~
    \begin{subfigure}[b]{1.0\textwidth}
        \subfigimg[trim={0cm 1.3cm 0cm 1cm}, clip=true, width=\textwidth, height=2cm]{Dist}{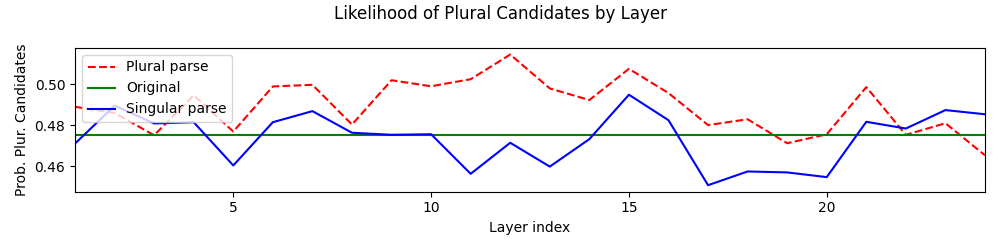}
    \end{subfigure}
    ~
    \begin{subfigure}[b]{1.0\textwidth}
        \subfigimg[trim={0cm 1.3cm 0cm 1cm}, clip=true, width=\textwidth, height=2cm]{2-layer Dist}{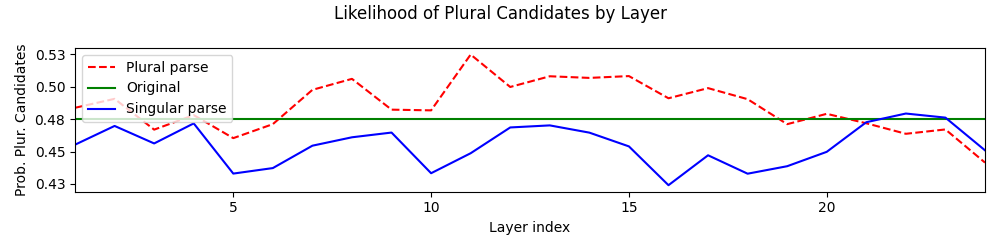}
    \end{subfigure}
    ~
    \begin{subfigure}[b]{\textwidth}
        \subfigimg[trim={0cm 0cm 0cm 1cm}, clip=true, width=\textwidth, height=2.8cm]{3-layer Dist}{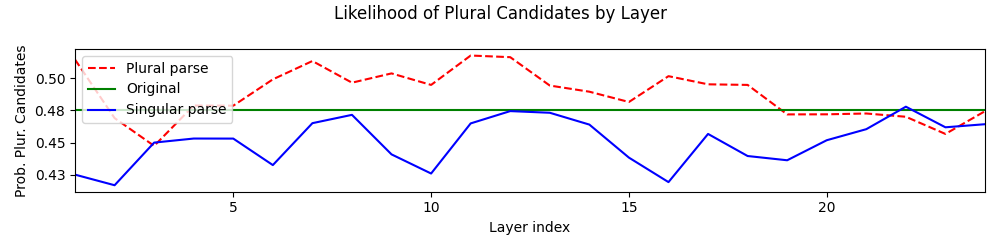}
    \end{subfigure}
    \caption{Mask model performance on the Coordination corpus. When using distance-based probes, the plural parse increased the likelihood of plural candidates being predicted, and the singular parse increased the likelihood of singular candidates being predicted.}
    \label{appfig:cloze_coord}
\end{figure*}

\begin{figure*}[!htb]
    \centering
    Mask Model Likelihood of Adverb Candidates by Layer in NP/Z Corpus
    \begin{subfigure}[b]{1.0\textwidth}
        \subfigimg[trim={0cm 1.3cm 0cm 1cm}, clip=true, width=\textwidth, height=2cm]{Depth}{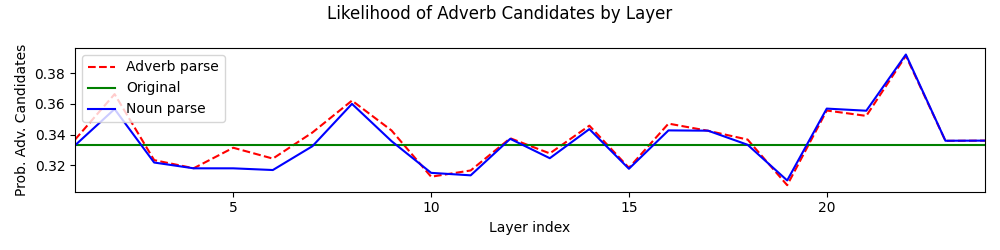}
    \end{subfigure}
    ~
    \begin{subfigure}[b]{1.0\textwidth}
        \subfigimg[trim={0cm 1.3cm 0cm 1cm}, clip=true, width=\textwidth, height=2cm]{Dist}{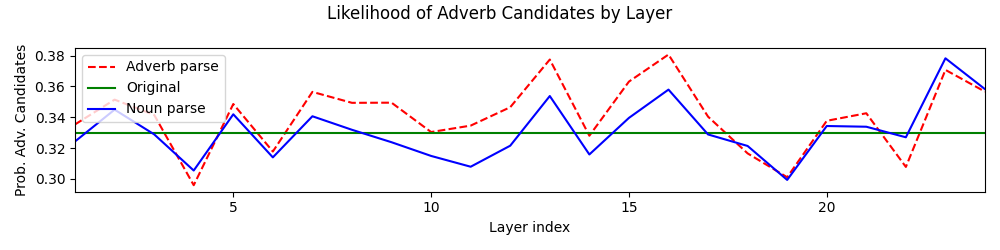}
    \end{subfigure}
    ~
    \begin{subfigure}[b]{1.0\textwidth}
        \subfigimg[trim={0cm 1.3cm 0cm 1cm}, clip=true, width=\textwidth, height=2cm]{2-layer Dist}{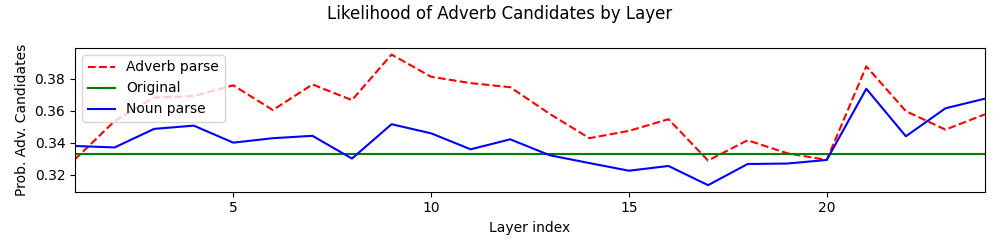}
    \end{subfigure}
    ~
    \begin{subfigure}[b]{\textwidth}
        \subfigimg[trim={0cm 0cm 0cm 1cm}, clip=true, width=\textwidth, height=2.8cm]{3-layer Dist}{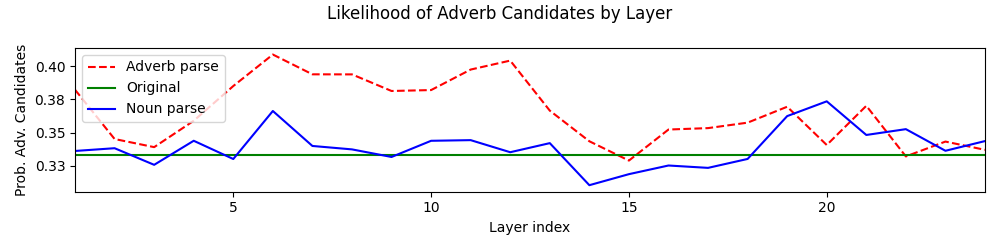}
    \end{subfigure}
    \caption{Mask model performance on the NP/Z corpus. Distance-based probes, and in particular multi-layer distance probes, changed model outputs according to syntactic principles.}
    \label{appfig:digging_merged}
\end{figure*}

\begin{figure*}[!htb]
    \centering
    QA Model Likelihood of NP1 Start by Layer in RC Corpus
    \begin{subfigure}[b]{1.0\textwidth}
        \subfigimg[trim={0cm 1.3cm 0cm 1cm}, clip=true, width=\textwidth, height=2cm]{Depth}{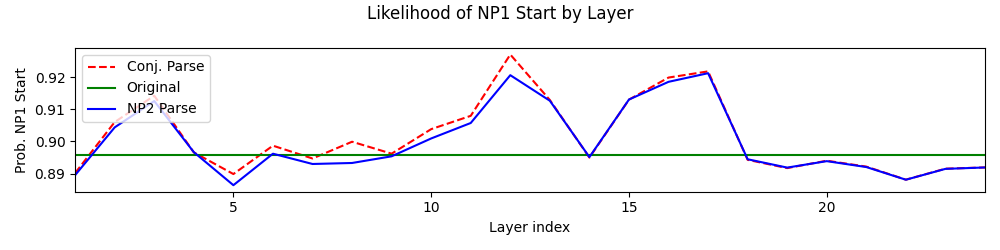}
    \end{subfigure}
    ~
    \begin{subfigure}[b]{1.0\textwidth}
        \subfigimg[trim={0cm 1.3cm 0cm 1cm}, clip=true, width=\textwidth, height=2cm]{Dist}{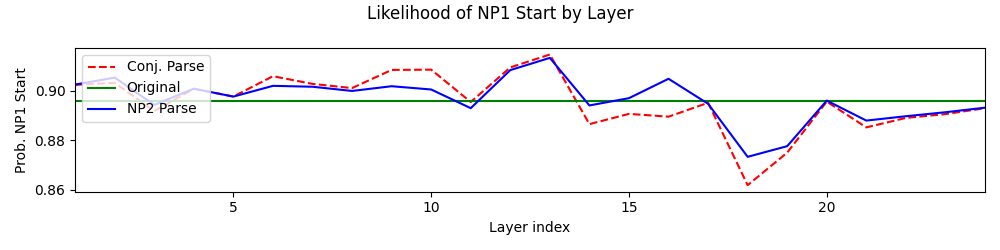}
    \end{subfigure}
    ~
    \begin{subfigure}[b]{1.0\textwidth}
        \subfigimgtop[trim={0cm 1.3cm 0cm 1cm}, clip=true, width=\textwidth, height=2cm]{2-layer Dist}{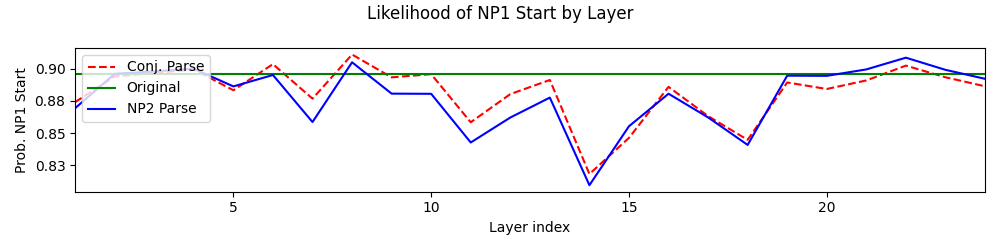}
    \end{subfigure}
    ~
    \begin{subfigure}[b]{\textwidth}
        \subfigimgtop[trim={0cm 0cm 0cm 1cm}, clip=true, width=\textwidth, height=2.8cm]{3-layer Dist}{supplementaryfigures/modifiers/modifiers_dist2.png}
    \end{subfigure}
    \caption{QA model performance on the RC corpus. No probe created consistent effects via counterfactual embeddings.}
    \label{appfig:qa_rc}
\end{figure*}

\begin{figure*}[!htb]
    \centering
    QA Model Likelihood of NP1 Start by Layer in NP/VP Corpus
    \begin{subfigure}[b]{1.0\textwidth}
        \subfigimgtop[trim={0cm 1.3cm 0cm 1cm}, clip=true, width=\textwidth, height=2cm]{Depth}{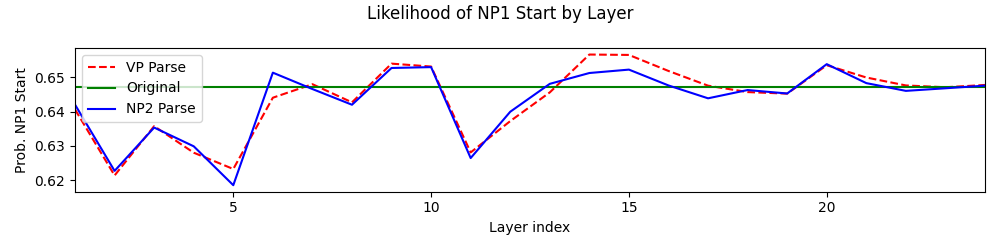}
    \end{subfigure}
    ~
    \begin{subfigure}[b]{1.0\textwidth}
        \subfigimg[trim={0cm 1.3cm 0cm 1cm}, clip=true, width=\textwidth, height=2cm]{Dist}{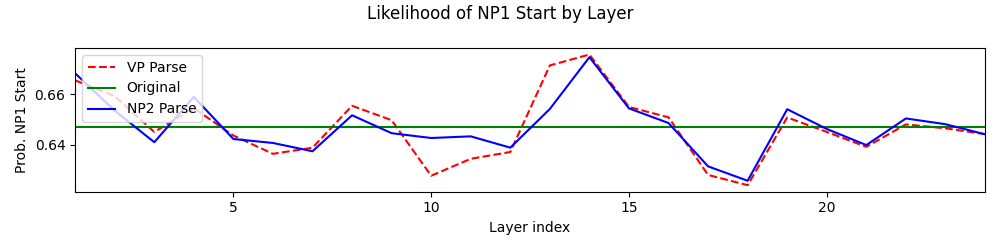}
    \end{subfigure}
    ~
    \begin{subfigure}[b]{1.0\textwidth}
        \subfigimg[trim={0cm 1.3cm 0cm 1cm}, clip=true, width=\textwidth, height=2cm]{2-layer Dist}{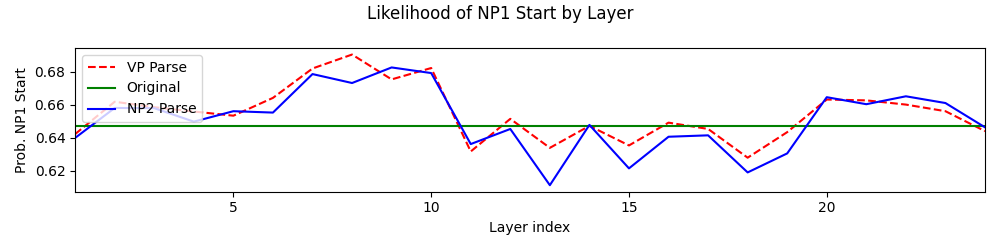}
    \end{subfigure}
    ~
    \begin{subfigure}[b]{\textwidth}
        \subfigimg[trim={0cm 0cm 0cm 1cm}, clip=true, width=\textwidth, height=2.8cm]{3-layer Dist}{supplementaryfigures/qa_np_vp/npvp_dist2.png}
    \end{subfigure}
    \caption{QA model on the NP/VP corpus. As in Figure~\ref{appfig:qa_rc}, no probe created consistent effects.}
    \label{appfig:qa_npvp}
\end{figure*}

In general, we observed small effects for counterfactuals in the QA Model (Figures~\ref{appfig:qa_rc} and \ref{appfig:qa_npvp}), but consistent effects in the Mask Model (Figures~\ref{appfig:cloze_coord} and \ref{appfig:digging_merged}).
Within the Mask model results, we also observed that the distance probe ($2^{nd}$ row) outperformed the depth probe ($1^{st}$ row), and that the multi-layer distance probes ($3^{rd}$ and $4^{th}$ rows) outperformed the linear distance probe.

\end{document}